\documentclass[letterpaper, 10 pt, conference]{ieeeconf}  

\IEEEoverridecommandlockouts                              

\title{\LARGE \bf
Improved Generalization of Heading Direction Estimation for Aerial Filming Using Semi-Supervised Regression 
}

\author{Wenshan Wang$^{1}$, Aayush Ahuja$^{1}$, Yanfu Zhang$^{2}$, Rogerio Bonatti$^{1}$ and Sebastian Scherer$^{1}$
\thanks{$^{1}$Wenshan Wang, Aayush Ahuja, Rogerio Bonatti and Sebastian Scherer are with the Robotics Institute
at Carnegie Mellon University, Pittsburgh, PA 15213, USA
        {\tt\small \{wenshanw,aahuja2,rbonatti,basti\}@andrew.cmu.edu}}%
\thanks{$^{2}$Yanfu Zhang is with Yamaha Motor Co., Ltd., Shizuoka 4370061, Japan
        {\tt\small zhangya@yamaha-motor.co.jp}}%
}

\usepackage{graphicx}
\usepackage{xcolor}
\usepackage{soul}
\usepackage{MnSymbol}
\usepackage[normalem]{ulem}
\usepackage{caption} 
\usepackage[font={small}]{caption}
\definecolor{remark}{rgb}{0, 0, 0}
\definecolor{highlight}{rgb}{0.9, 0.17, 0.31}
\begin{document}

\maketitle
\thispagestyle{empty}
\pagestyle{empty}

\begin{abstract}
In the task of Autonomous aerial filming of a moving actor (e.g. a person or a vehicle), it is crucial to have a good heading direction estimation for the actor from the visual input. However, the models obtained in other similar tasks, such as pedestrian collision risk analysis and human-robot interaction, are very difficult to generalize to the aerial filming task, because of the difference in data distributions.   
Towards improving generalization with less amount of labeled data, this paper presents a semi-supervised algorithm for heading direction estimation problem. We utilize temporal continuity as the unsupervised signal to regularize the model and achieve better generalization ability. This semi-supervised algorithm is applied to both training and testing phases, which increases the testing performance by a large margin. 
We show that by leveraging unlabeled sequences, the amount of labeled data required can be significantly reduced. We also discuss several important details on improving the performance by balancing labeled and unlabeled loss, and making good combinations. Experimental results show that our approach robustly outputs the heading direction for different types of actor. The aesthetic value of the video is also improved in the aerial filming task.  

\end{abstract}

\section{INTRODUCTION}

Aerial filming is popular with both professional and amateur film makers due to its capability of composing viewpoints that are not feasible using traditional hand-held cameras. With recent advances in state estimation and control technology for multi-rotor vehicles, consumer-level drones became easier to operate; even amateur pilots can easily use them for static landscape filming. However, aerial filming for moving actors in action scenes is still a difficult task for drone pilots and requires considerable expertise. There is increasing demand for an autonomous cinematographer; one that can track the actor and capture exciting moments without demanding attention and effort from a human operator \cite{bonatti2018}.


When filming a moving actor with an autonomous drone, heading direction estimation (HDE) plays a central role, both in terms of motion planning and scene aesthetics. With accurate heading information, the planner can make a better forecast of the actor's future movement, enabling smoother motion, and more reliable visual tracking. In addition, by knowing the actor's heading the drone can execute different types of shots  (\textit{e.g} front, back, side shots), depending on the user's artistic objectives.

\begin{figure}[h]
    \centering
    \includegraphics[width=0.45\textwidth]{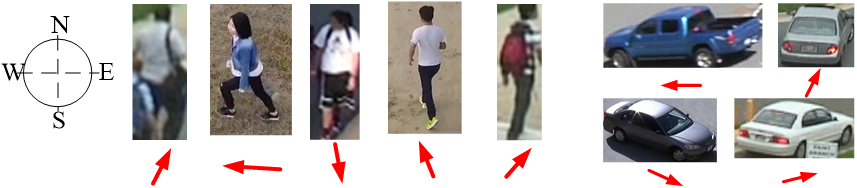}
    \caption{The HDE task. Given the images, we want to predict the heading direction as shown by the red arrows.}
    \label{fig:heading_problem}
\end{figure}



Estimating the heading of people and objects is also a problem within many other applications, such as pedestrian collision risk analysis \cite{tian2014estimation}, human-robot interaction \cite{vazquez2015parallel} and activity forecasting \cite{kitani2012activity}. However, models obtained for other tasks do not easily generalize to the aerial filming task, due to a mismatch in the types of images from datasets coming from each application. Aerial images tend to have large variations in terms of angles, scale, brightness, and blurriness of the actor (Figure~\ref{fig:dataset}). In addition, when the trained model is deployed on a drone, the input actor images used for the HDE module result from an imperfect object detection module, increasing the mismatch between existing datasets \cite{ristani2016MTMC}\cite{Geiger2013IJRR} and the data seen in practice.




Despite the success of deep learning methods, the generalization problem is always one of the major stumbling blocks for applying those methods to real robotic problems, since they rely heavily on labeled data. This is a problem not only because labeling is a laborious task, but also because the labeled data could be biased, resulting in a poor generalization to other applications.
Since there are many different types of filming targets such as persons and vehicles,  decreasing the amount of labeled data required could be extremely helpful for applying the methods across object types.



This paper presents a semi-supervised algorithm for HDE problem. We utilize temporal continuity as the unsupervised signal to regularize the model and achieve better generalization ability. This semi-supervised algorithm is applied to both training and testing phases, which increase the testing performance by a large margin. 
We show that by leveraging unlabeled sequences, the amount of labeled data required can be significantly reduced. We also discuss several important details on improving the performance by balancing labeled and unlabeled loss. Experimental results show that our approach robustly outputs the heading direction for different actors in filming task. This results in a better state estimation for the actor in 3D space, also allows us to obtain more visually appealing videos. 
\begin{figure*}[h]
    \centering
    \includegraphics[width=0.95\textwidth]{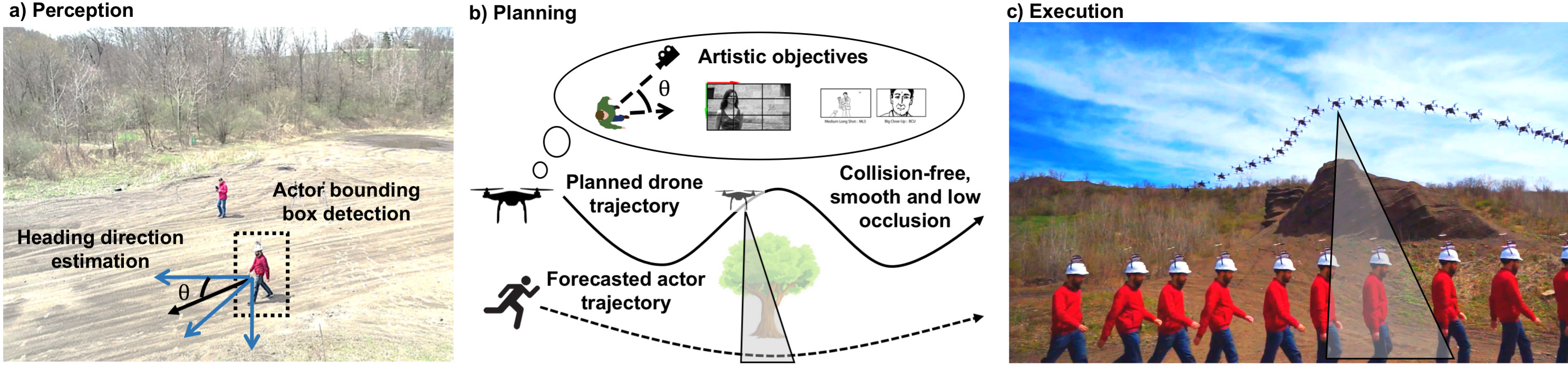}
    \caption{We propose a system for automatic heading direction estimation (HDE) of a subject, applied for autonomous aerial cinematography. a) The perception system detects a bounding box around the actor, and the HDE network outputs the current heading angle in the world frame. b) The drone reasons about artistic guidelines provided by a user including relative angle to the actor, and calculates a smooth, collision-free trajectory, avoiding occlusion. c) Real-life results validate the system working on an on-board computer, following and tracking the subject online.}
    \label{fig:cover_pic}
\end{figure*}

In this work we offer three main contributions: 1) We propose a novel semi-supervised approach that uses temporal continuity in sequential data for heading direction estimation problem; 2) We compare our method with baseline approaches, showing that we significantly reduce the amount of labeled data required to train a robust model; and 3) We experimentally verify the robustness of our proposed method, while running on an onboard computer of a drone, following an actor.

\section{RELATED WORK}

Heading direction estimation is a widely studied problem, in particular focused on humans and cars. One option to tackle the problem is to use inertial and GPS sensors to estimate human's \cite{liu2016novel,deng2017heading} or a car's \cite{vista2015design} heading direction. In the context of aerial filming, the target actor generally does not carry extra sensors; thus our emphasis on vision-based solutions in this paper. 

Based on a probabilistic framework, Flohr et al. \cite{flohr2015probabilistic} present a joint pedestrian head and body orientation estimation method, in which they design a HOG/linSVM pedestrian detector combined with a Kalman filter. 
Learning-based methods, however, seem to achieve more robust and generalizable results, being more prevalent in the HDE literature.
Most existing learning-based methods use large amounts of labeled data and supervised learning to train a model 
\cite{liu2017weighted,choi2016human,braun2016pose,prokudin2018deep}.
However, open datasets \cite{h36m_pami, raman2016direction, liu2013accurate, Geiger2013IJRR} generalize poorly to our aerial filming task mainly due to mismatch between image viewpoints, scales, and image blurr.
Human key-point detection and 2D pose estimation have also been widely studied \cite{toshev2014deeppose, cao2017realtime}. However, such works are focused only on human bodies, and the 3D heading direction cannot be trivially recovered directly from 2D points because the keypoint's depth remains undefined.


Semi-supervised learning (SSL) is also an active research area. Self-training is a commonly used technique for SSL \cite{yarowsky1995unsupervised}. Weston et al. proposed a graph-based method for semi-supervised classification \cite{weston2012deep}, and recently more related works have been proposed  \cite{hoffer2016semi,rasmus2015semi,dai2017good} in the area. Most of the existing SSL works are focused on classification problems, which assume that different classes are separated by a low-density area. 
This assumption is not directly applicable to regression problems.


Temporal continuity is a valid assumption for many robotics sensory data and can be exploited to improve the training of neural networks using SSL. 
Mobahi et al. \cite{mobahi2009deep} developed one of the first approaches to exploit the temporal continuity with deep convolutional neural networks. The authors use video temporal continuity as a pseudo-supervisory signal over the unlabeled data and demonstrate that this additional signal can improve object recognition in videos from 
the COIL-100 dataset \cite{nene1996columbia}. Other works learn feature representations by exploiting temporal continuity \cite{zou2012deep, goroshin2015unsupervised, stavens2010unsupervised, srivastava2015unsupervised, wang2015unsupervised}. Zou et al. \cite{zou2012deep} included the video temporal constraints in an autoencoder framework and learned invariant features accross frames. Wang et al.\cite{wang2015unsupervised} designed a Siamese-triplet network which can be trained in an unsupervised manner with a large amount of video data. Wang et al. showed that the unsupervised visual representation can achieve competitive performance on various tasks, compared to its ImageNet-supervised counterpart. Srivastava et al. \cite{srivastava2015unsupervised} trained a LSTM model in an unsupervised manner by exploiting temporal continuity and demonstrated that this learned representation of video sequences could be used for predicting future frames or improving action recognition accuracy. 

Based on the concept of leveraging temporal continuity, we aim to improve the learning of a regression model from a small labeled dataset. The small dataset constraint is common in many robotics applications, where the cost of acquiring and labeling more data is high. 
We describe our approach in the following section. 

\section{APPROACH}

\subsection{Problem Formulation} 

To estimate the actor's position and orientation in the world frame, we define HDE and ray-casting modules that process the incoming camera image.
We define the pose of the actor as a vector $[x, y, \theta_w]$ on the ground plane. The actor's position $[x, y]$ is inferred by the ray-casting module; it detects a bounding box around the actor at each camera frame, and projects the center of the box's bottom onto the ground plane using the known extrinsic and intrinsic camera parameters. To estimate the $\theta_w$ component, we first estimate the actor's heading $\theta$ from the image within the bounding box (Figure~\ref{fig:heading_problem}), and then convert $\theta$ to the world frame coordinates using the camera parameters. 

The HDE module works by outputting the angle $\theta$ in image space. Since $\theta$ is ambiguously defined at the frontier between $-\pi$ and $\pi$, we define the output of the regressor as two continuous values $[cos(\theta), sin(\theta)]$, therefore avoiding instabilities in the model during training and inference. 

We assume we have access to a relatively small labeled dataset $D=\{(x_i, y_i)\}_{i=0}^{n}$, where $x_i$ denotes input image, and $y_i=[cos(\theta_i), sin(\theta_i)]$ denotes the angle label. In addition, we assume access to a large unlabeled sequential dataset $U=\{q_j\}_{j=0}^{m}$, where $q_j=\{x_0, x_1,...,x_t\}$ is a sequence of temporally-continuous image data. The HDE module's objective is to approximate a function $y=f(x)$, that minimizes the regression loss on the labeled data $ \sum_{(x,y)\in{D}}L_l(x_l,y_l) = \sum_{(x,y)\in{D}}||y_i-f(x_i)||^2$. 


\subsection{Defining the loss for temporal continuity}

One intuitive way to leverage unlabeled data is to add a constraint that the output of the model should be continuous over a consecutive input sequence. In this sense, the model is trained to jointly minimize the labeled loss $L_l$ and the continuity loss $L_u$. We minimize the combined loss:  
$$ \min{\sum _{(x_l,y_l)\in{L}}L_l(x_l,y_l)+\lambda \sum_{q_u\in{U}}{L_u(q_u)}} \eqno{(1)}$$

The labeled loss could be the mean squared error on labeled data. The continuity loss could be defined in many ways. Intuitively, we look at the samples within one sequence. If two samples are close with respect to temporal distance, their outputs should also be close. 
$$ L_u(q_u) = \sum_{x_1\in{q_u},x_2\in{q_u}} S(x_1,x_2) D(x_1,x_2;f)  \eqno{(2)}$$
where $S$ is the similarity between two inputs measured by temporal distance, $D$ is the difference between two outputs produced by the network. To minimize the continuity loss, if two samples within the same sequence are close to each other, their output difference should be small. 

The similarity could be defined as follows which takes into account the temporal distance (i.e. the number of consecutive points between the two given points) and it's decay over time. 
$$ S(x_1,x_2) = e^{-\alpha|n_{x_1}-n_{x_2}|}  \eqno{(3)}$$

where $\alpha$ controls the decay speed, $n_x$ is the frame number of input $x$. The output difference between two samples is taken as the Euclidean distance of the output layer. 
$$ D(x_1,x_2;f) = ||f(x_1)-f(x_2)||_2 \eqno{(4)}$$

In practice, we add a small threshold that allows a small difference between consecutive samples. 
The problem with the previous formulation is the introduction of an additional hyper-parameter $\alpha$. The hyper-parameter needs to be tuned for different datasets which is difficult and time-consuming in practise. Instead, we look for alternate formulations that alleviate the use of additional hyper-parameter and relax this criterion.

The following formulation of the unsupervised loss is based on the idea that close samples should output smaller difference than far away samples. Similar continuity loss is also used in \cite{wang2015unsupervised} when training an unsupervised feature extractor. 
$$ L_u(q_u) = \sum_{x_1,x_2,x_3} \max[0,D(x_1,x_2;f)-D(x_1,x_3;f)]  \eqno{(6)}$$
where $x_1,x_2,x_3\in{q_u}$ and $S(x_1,x_2)>S(x_1,x_3)$. This usually works better than the previous loss because it doesn't contain additional hyper-parameters and extracts more information from the unlabeled data. The loss definition in Equation 2 indicates two temporally close inputs should have similar outputs. While the loss definition in Equation 6 also suggests that farther inputs should have bigger output difference than closer inputs. This could potentially decrease the small oscillation in output sequence. 




\subsection{Network structure}

\begin{figure}[h]
    \centering
    \includegraphics[width=0.5\textwidth]{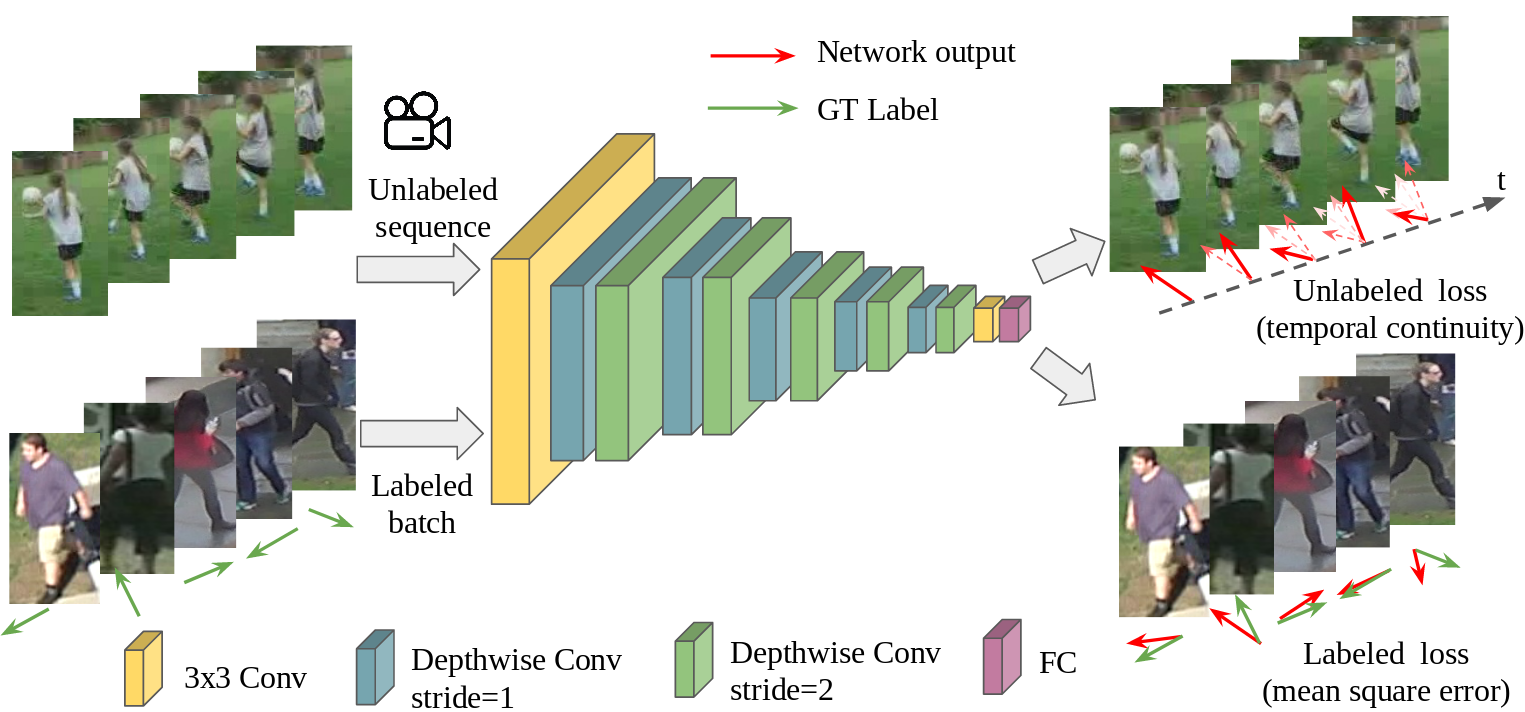}
    \caption{Our architecture for predicting heading direction. We use mobilenet-based feature extractor followed by a convolutional layer and a fully connected layer to predict angular values. The network is trained using both labeled and continuity loss.}
    \label{fig:heading_structure}
\end{figure}

In order to use our network on an onboard computer for real-time applications,
we utilize a compact convolutional neural network based on MobileNet \cite{howard2017mobilenets}. The input of the network is a cropped image of the target, outputted by the detection and tracking modules. The cropped image is padded to a square shape and resized to 192 x 192 pixels. After 10 group-wise and point-wise convolutional blocks as described by MobileNet, we add another convolutional layer and a fully connected layer that output two values, representing the cosine and sine values of the angle (Figure~\ref{fig:heading_structure}). 

During each training iteration, one shuffled batch of labeled data and one sequence of unlabeled data are passed through the network. The labeled loss and the unlabeled loss are computed and backpropagated through the network.


\subsection{Cross-dataset semi-supervised fine-tuning}

We use the network architecture explained in the previous section to train on openly accessible datasets. 
Due to the significant differences in the data distribution between openly available datasets and our drone filming data, the average angle error of the network when tested on the drone filming data is above 30 degree, which is still not good enough for real-world application. To address the cross-dataset generalization problem, our key insight is to apply the same semi-supervised training idea to finetune the model by employing a few labelled and unlabelled sequences from the drone filming data.

In our drone filming task, the input data being a video is naturally available in a sequential manner, thus allowing the use of the unsupervised loss. We use the insight developed earlier and label a few samples from the video data collected. The HDE model pretrained on openly accessible datasets is finetuned with both labeled loss and unlabeled loss on this collected data. As we will show in the experiment section, using a semi-supervised approach, our model generalizes much better to our drone filming task on unseen sequences and images. Compared with employing only the supervised loss on the drone filming data, it will be shown that generalization is improved while achieving robust and stable performance by employing the unsupervised loss. Our approach makes it feasible for application to other datasets or different kinds of objects since few labelled examples are required and collecting unlabelled videos is easier.

\section{EXPERIMENTS}

\subsection{Dataset and baselines}

We collected a large number of image sequences from various sources. For the person HDE, we use two surveillance datasets: VIRAT \cite{oh2011large} and DukeMCMT~\cite{ristani2016MTMC}, and one action classification dataset: UCF101 \cite{soomro2012ucf101}. However, none of these datasets provides ground truth for HDE. Therefore, in the UCF101 dataset, we manually labeled 453 images for HDE. In the surveillance datasets, we adopted a semi-automatic labelling approach that we first detect the actor in every frame, then compute the ground-truth heading direction based on the derivative of the subject's position over a sequence of consecutive time frames. For the car HDE, two surveillance datasets VIRAT and Ko-PER~\cite{strigel2014ko} and one driving dataset PKU-POSS \cite{wang2016probabilistic} is utilized (Table~\ref{heading_datasets}). 

We find the HDE problem for cars is easier because they are more rigid than human. We use humans as example for the rest of the paper. 

\begin{table}[h]
\caption{Datasets used in this study}
\label{heading_datasets}
\begin{center}
\begin{tabular}{p{0.8cm} p{0.9cm} p{0.9cm} p{1.2cm} p{0.8cm} p{0.9cm}}
\hline
dataset & VIRAT & UCF101 & DukeMCMT & Ko-PER & PKU-POSS \\
\hline
target & car/person & person   & person & car       & car \\
GT      & MT*       & HL(453)* & MT*    & $\checkmark$ & $\checkmark$ \\
\# seqs & 650       & 940      & 4336   & 12         & - \\
\# imgs & 69680     & 118027   & 274313 & 18277     & 28973 \\
\hline
\end{tabular}
\end{center}
      \small
      *MT denotes labeling by motion tracking, HL denotes manual labeling.
\end{table}

\begin{figure}[h]
    \centering
    \includegraphics[width=0.4\textwidth]{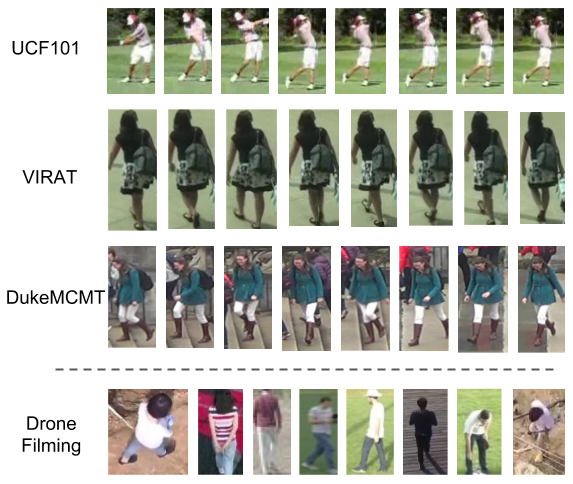}
    \caption{The data for human HDE problem. The data distribution of aerial filming task is very different from the distribution of the open accessible datasets, in terms of image size, blurriness, shot angle and human pose. }
    \label{fig:dataset}
\end{figure}

The data distribution of these datasets is quite different from the data distribution of the drone filming task. The problem is quite challenging since most of the images in our task are very small and blurred (Figure~\ref{fig:dataset}). 

We compare our approach against two baselines for HDE. The first baseline Vanilla-CNN is a simple convolutional neural network inspired by \cite{choi2016human}. The second baseline CNN-GRU implicitly learns temporal continuity using a GRU network inspired by \cite{liu2017weighted}. One drawback for this model is that although it models the temporal continuity implicitly, it needs large number of labeled sequential data for training, which is very expensive to obtain. 

We employ three metrics for quantitative evaluation: 1) Mean square error (MSE) between the output $(\cos\theta, \sin\theta)$ and the ground truth $(\hat{\cos}\theta, \hat{\sin}\theta)$. 2) Angular difference (AngleDiff) between the output and the ground truth. 3) Accuracy obtained by counting the percentage of correct outputs, which satisfies AngleDiff$<\pi/8$. We use the third metric, which allows small error, to alleviate the ambiguity in labeling human heading direction.

\subsection{Deceasing labeled data using semi-supervised regression}

\begin{figure}[h]
    \centering
    \includegraphics[width=0.5\textwidth]{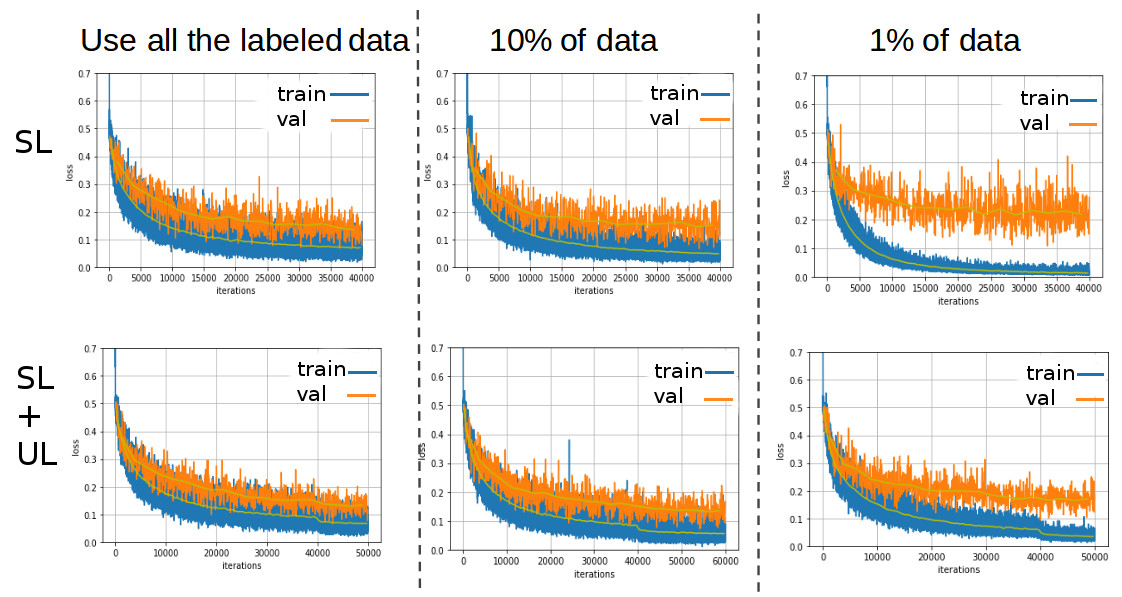}
    \caption{The top row shows training and validation loss for supervised learning using different number of labeled data. The validation performance drops from 0.13 to 0.22, when decreasing the number of labeled data from 100\% to 1\%. The bottom row shows results with semi-supervised learning. The validation losses are 0.13, 0.14 and 0.17 respectively for 100\%, 10\% and 1\% labeled data. }
    \label{fig:semi_compare}
\end{figure}

In this experiment, we train the regression network on the DukeMCMT dataset, which consists of 274k labeled images. Those images are taken from 8 different surveillance cameras. We use the data from 7 of them for training, and one of them for testing (about 50k). We compared our semi-supervised method with supervised one using different number of labeled data and the result is shown in Figure~\ref{fig:semi_compare}. We verify that by utilizing unsupervised loss as Equation (2), the model generalizes better to the validation data than the one with purely supervised loss. 

\subsection{Improve generalization ability with semi-supervised finetuning}

We first train a HDE network on datasets shown in Table~\ref{heading_datasets}. Those models perform well on the validation set from those datasets. However they generalize poorly to the drone filming task, as shown in row 1-3 of Table~\ref{tab:semi_finetune}. It is necessary to finetune the model on our data. To minimize the labeling effort, and improve the generalization capbility, we employ semi-supervised method to the finetuning process. 

\begin{figure}[h]
    \centering
    \includegraphics[width=0.45\textwidth]{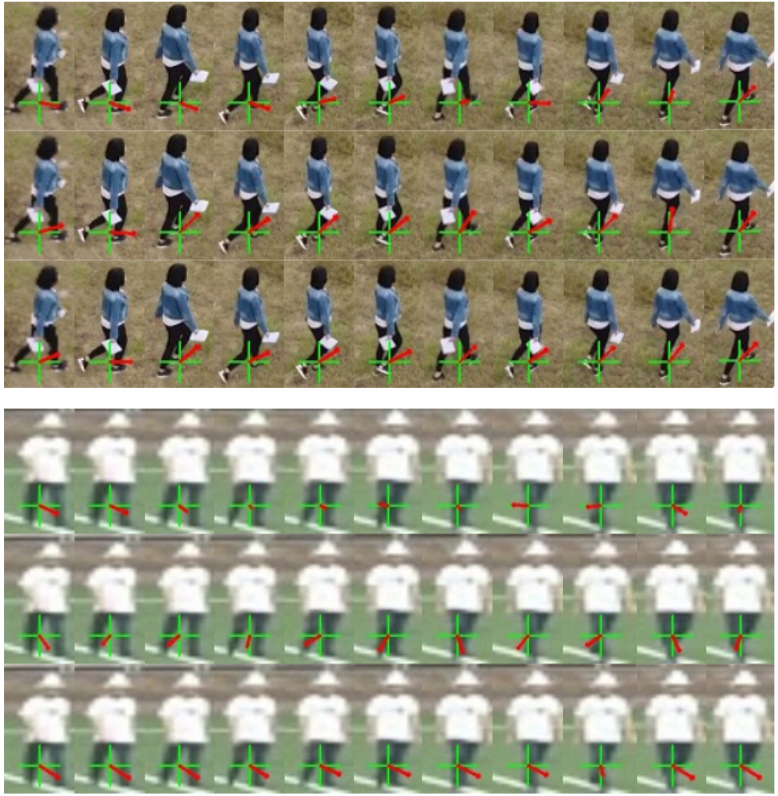}
    \caption{Three models are tested on the sequential data. Two testing sequences are shown in this figure. The top row of each testing sequence shows the results that directly employ the model trained on other open accessible datasets to the aerial filming task. It generalizes poorly due to the distribution difference. The middle row and bottom row show the results after finetuning the model on the filming data with and without continuity loss respectively. The model using continuity loss for finetuning (bottom row) outputs more accurate and smooth results. }
    \label{fig:comp_finetune}
\end{figure}

In our aerial filming task, the input data is naturally in consecutive manner.  We collect around 50 videos, each contains approximately 500 sequential images. For each video, we manually labeled 6 of those images. The HDE model is finetuned with both labeled loss and continuity loss, same as the training process on the open accessible datasets. 
We qualitatively and quantitatively show the results of HDE using semi-supervised finetuning in Figure~\ref{fig:comp_finetune} and Table~\ref{tab:semi_finetune}. The experiment verifies our model could generalize very well to our drone filming task. Compared with the purely supervised learning approach, utilizing unlabeled data improves generalization results and achieves more robust and more stable performance.

We use three metrics to evaluate different methods on aerial filming task (Table~\ref{tab:semi_finetune}). Vanilla-CNN and CNN-GRU baselines trained on open datasets don't transfer well to drone filming dataset with accuracy below 30\%. Our SSL based model trained on open datasets achieves 48.7\% accuracy. By finetuning on labelled samples of drone filming, we improve this to 68.1\%. Best performance is achieved by finetuning on labelled and  unlabeled sequences of the drone filming data with accuracy 72.2\%.

\begin{table}[h]
\caption{Semi-Supervised Finetuning results}
\label{tab:semi_finetune}
\begin{center}
\begin{tabular}{c p{1.1cm} p{1.1cm} p{1.1cm}}
\hline
Method & MSE loss & AngleDiff (rad) & Accuracy (\%) \\
\hline 
Vanilla-CNN \cite{choi2016human} w/o finetune & 0.53 & 1.12 & 26.67\\
CNN-GRU \cite{liu2017weighted} w/o finetune & 0.5 & 1.05 & 29.33 \\
SSL w/o finetune & 0.245 & 0.649 & 48.7  \\
SL w finetune  & 0.146 & 0.370 & 68.1 \\
SSL w finetune  & \textbf{0.113} & \textbf{0.359} & \textbf{72.2} \\
\hline
\end{tabular}
\end{center}
\end{table}

\subsection{Implementation details}

The $\lambda$ in Equation (1) is an important hyper-parameter to balance the two losses. These two losses have separate objectives. When $\lambda$ is small, the continuity loss decreases slowly, while the gap between the training loss and validation loss remains substantial. With large $\lambda$, the continuity loss is very low, while both training and validation loss decrease slowly. The network tend to output trivial unchanged outputs to keep the continuity loss low. We achieve best performance using a semi-supervised approach with $\lambda=0.1$. 


Since collecting unlabeled data is easy, the question is whether having more unlabeled data is always better? If it were true, we could simply collect more data and solve most of the regression problems. Unfortunately, this is not the case. We observe that the unlabeled data improves the performance most only when it is from a similar distribution as the labeled data. 

Intuitively speaking, if the distribution of unlabeled data is far from the labeled data, the network would have no idea on the true values of the sequential frames, and it would output continuous values for the sequence in a arbitrary way. To verify this, we evaluated the effect of adding new data in four different settings, and show the validation results on filming data. 

1) Add labeled data only: 500 labeled samples (filming dataset).

2) Add unlabeled data only: 1485 unlabeled sequences (filming dataset).

3) Labeled + unlabeled (same distribution): 500 labeled (filming dataset) + 1485 unlabeled sequences (filming dataset).

4) Labeled + unlabeled (different distribution): 500 labeled (UCF101 dataset) + 1485 unlabeled sequences (filming dataset).

\begin{figure}[h]
    \centering
    \includegraphics[width=0.4\textwidth]{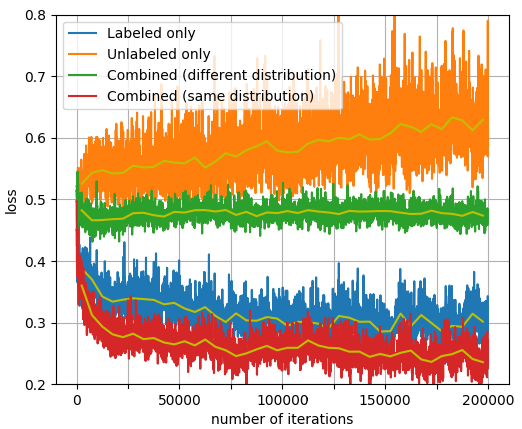}
    \caption{Validation loss curves on filming data, under four different settings of adding unlabeled data. Orange curve represents adding only unlabeled filming data. Without labeled signal, the validation performance get worse. The green one shows the result of adding labeled and unlabeled data from different distribution. The validation loss drops at first but stay high. The blue curve shows adding only the labeled filming data, which performs well. The red curve shows best result when using both labeled and unlabeled filming data.}
    \label{fig:distribution_cmp}
\end{figure}

The results of these four settings are shown in Figure~\ref{fig:distribution_cmp}, which plots the validation loss on the filming data. The validation result achieves the best performance when including both the labeled and the unlabeled data from the filming dataset. This answers the question that the unlabeled sequences alone does not increase the performance. It is important to add a small number of labeled data from similar distribution with the unlabeled sequences.

\subsection{Drone Cinematography using HDE Results}

We applied the HDE system to an aerial cinematography platform, and evaluated the real-time HDE results, in addition to improvements in filming quality in comparison with baseline methods for heading estimation. In general, the aerial cinematography platform is broadly divided into vision and planning subsystems. In the vision subsystem, we first detect and track the actor using a monocular camera. We use the bounding box of the tracked actor as an image-space signal to control the camera gimbal, keeping it in the desired screen position. We further adopt the HDE component and a ray-casting approach to estimate the actor's current pose, and a learning-based method~\cite{irl} to forecast the actor's future poses in the world frame. Details about the hardware and planning subsystem can be found in the work of Bonatti et al. \cite{bonatti2018}. 

\begin{figure}[h]
    \centering
    \includegraphics[width=0.5\textwidth]{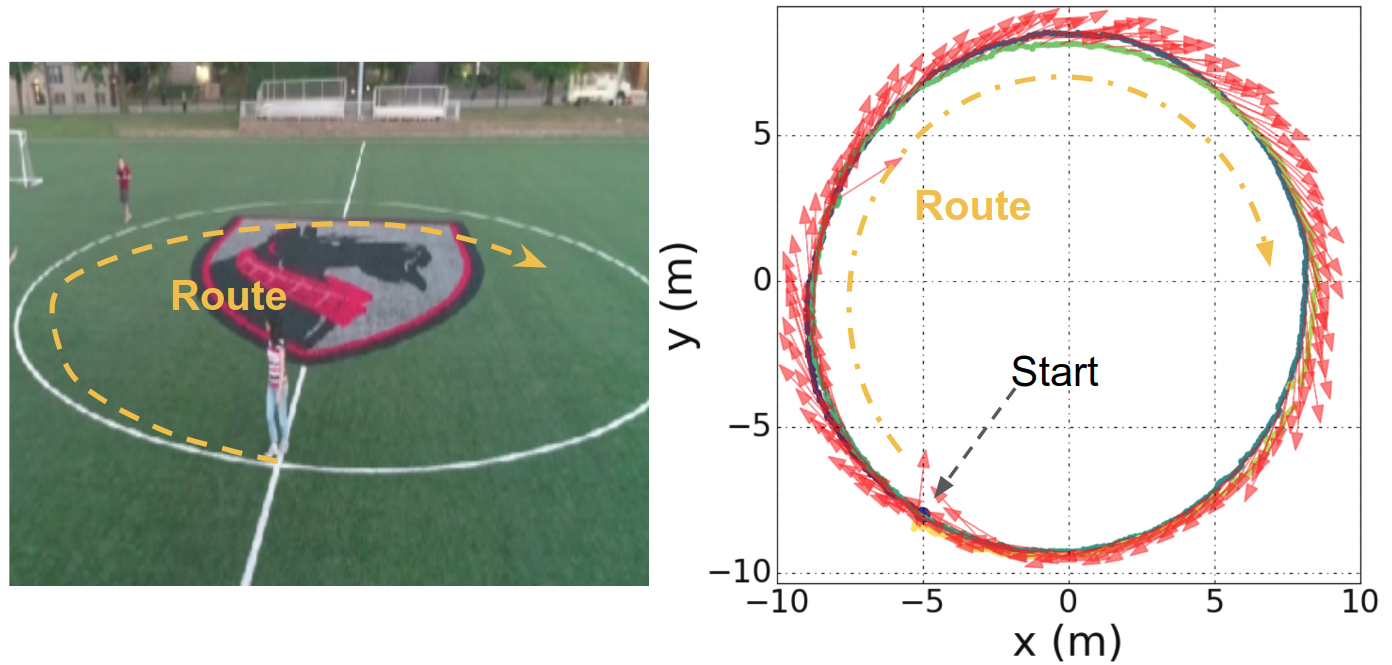}
    \caption{Qualitative evaluation of the HDE results. The actor walks twice on a circle as shown in the left figure. The actor intentionally keeps the heading to be the same as the tangential direction w.r.t. the circle. In the right figure, red arrows represent the estimated actor heading direction. We can see that the estimated heading direction keeps good consistence and aligns well with tangential direction. With the accurate heading estimation, various shot types, such as side shot or front shot, can be achieved without requiring the actor to wear additional sensor setup.}
    \label{fig:ray_casting}
\end{figure}

\begin{figure}[h]
    \centering
    \includegraphics[width=0.4\textwidth]{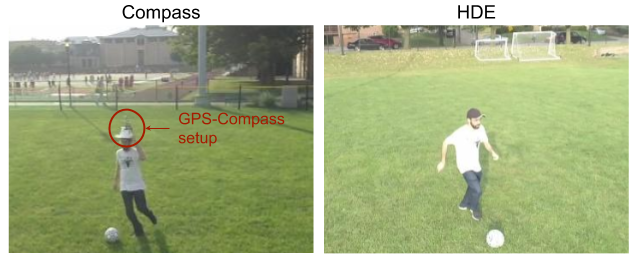}
    \caption{Comparison of resulting drone images taken using a compass-GPS setup and visual HDE. The use of a compass and GPS module significantly decrease image aesthetics, and proves to be impractical in most real-life applications for a drone cinematographer. Meanwhile, the visual HDE requires less hardware, is simpler, and allows more complex and constraint-free actor motions.}
    \label{fig:real_life}
\end{figure}

Due to the limited ways to obtain accurate ground truth in an outdoor environment, we qualitatively evaluate the HDE results as shown in Figure~\ref{fig:ray_casting}. We can see that the HDE system gives consistent and accurate estimations. Before adopting the HDE system, we use an external GPS/compass setup to obtain the actor's pose. Then applying motion forecasting and planning techniques to achieve pre-defined filming patterns. However, this external sensor setup is highly undesired because it significantly constraints the actor's motion and degrades the image aesthetics. In the complementary video, we show that the HDE system can be integrated into the aerial cinematography system and replace the external sensor setup. And Figure~\ref{fig:real_life} shows that replacing the external sensor setup with the HDE system can greatly improve the aesthetics of the images captured by the drone cinematographer.

\section{CONCLUSIONS}


In this work, we propose a semi-supervised method for HDE problems by leveraging temporal continuity in consecutive unlabeled inputs. 
We employ the  semi-supervised framework to training and finetuning steps, and show that it significantly reduce the amount of labeled data required to train a robust model. We experimentally verify the robustness of our proposed method, while running on an onboard computer of a drone, following an actor. We plan to apply this framework to other mobile robot tasks in the future.






\section*{ACKNOWLEDGMENT}

Research presented in this paper was funded by Yamaha Motor Co., Ltd. under award \#A019969 .



\bibliographystyle{ieeetr}
\bibliography{ref.bbl}

\end{document}